\pdfoutput=1

\documentclass[11pt]{article}

\usepackage[preprint]{acl}
\usepackage{tabularx}

\usepackage{times}
\usepackage{latexsym}
\usepackage{multirow}
\usepackage{authblk}
\usepackage[T1]{fontenc}

\usepackage[utf8]{inputenc}

\usepackage{microtype}

\usepackage{inconsolata}

\usepackage{graphicx}

\title{Enhancing Document-level Argument Extraction with Definition-augmented Heuristic-driven Prompting for LLMs}

\author[1]{Tongyue Sun}
\author[2,3]{Jiayi Xiao}
\affil[1]{School of Engineering and Informatics,
    University of Sussex,
    Brighton, UK}
\affil[2]{International Business School Suzhou, Xi’an Jiaotong-Liverpool University, Suzhou, China}
\affil[3]{Management School, University of Liverpool, Liverpool, UK}

\begin{document}
\maketitle
\begin{abstract}
Event Argument Extraction (EAE) is pivotal for extracting structured information from unstructured text, yet it remains challenging due to the complexity of real-world document-level EAE. We propose a novel Definition-augmented Heuristic-driven Prompting (DHP) method to enhance the performance of Large Language Models (LLMs) in document-level EAE. Our method integrates argument extraction-related definitions and heuristic rules to guide the extraction process, reducing error propagation and improving task accuracy. We also employ the Chain-of-Thought (CoT) method to simulate human reasoning, breaking down complex problems into manageable sub-problems. 
Experiments have shown that our method achieves a certain improvement in performance over existing prompting methods and few-shot supervised learning on document-level EAE datasets. The DHP method enhances the generalization capability of LLMs and reduces reliance on large annotated datasets, offering a novel research perspective for document-level EAE.

\end{abstract}

\section{Introduction}

Event Argument Extraction (EAE) is a task within the domain of Natural Language Processing (NLP), focusing on the identification of relevant information pertaining to specific events from textual data. 
The majority of previous studies posit that events are articulated solely within a single sentence, hence their primary focus has been on sentence-level information extraction \cite{Chen_Xu_Liu_Zeng_Zhao_2015, liu-etal-2018-jointly, zhou2021role}. 
However, in real-life contexts, events are often narrated through complete documents composed of multiple sentences, such as news reports or medical records, an area that remains to be thoroughly investigated.
Document-level EAE commonly relies on manual domain and pattern annotation for supervised learning models \cite{xiang2019survey, Lin_Ji_Huang_Wu_2020, li2022survey, liu-etal-2022-dynamic, hsu2022degree, liu2023document}. While this method is effective, it requires substantial labeling work. Considering the inherent complexity of document-level EAE, this is particularly burdensome and costly.

With the continuous evolution of Large Language Models (LLMs), their demonstrated potential has positioned them as formidable competitors to traditional methods in the field of EAE. For instance, InstructGPT \cite{ouyang2022training} and ChatGLM \cite{du2022glm} have excelled in diverse downstream applications such as dialogue systems and text summarization generation through meticulously crafted instructions. Furthermore, recent studies \cite{lin-etal-2023-global, zhou2024llms} have expanded the application of LLMs in complex tasks like event extraction by ingeniously constructing prompts, highlighting the broad prospects of LLMs in the EAE domain. 

In prior research, pre-trained and fine-tuned models have exhibited deficiencies in generalization capabilities, largely constrained by the high costs of annotation and the risks of error propagation. The domain of document-level event argument extraction faces significant challenges, with the scarcity of high-quality datasets and the models' insufficient generalization to unseen events being the primary bottlenecks.
In contrast to the traditional reliance on vast corpora, the incorporation of In-Context Learning (ICL) within LLMs has emerged as a transformative approach \cite{brown2020language, zhou2022least, zhou2023heuristics, wang2024large}. 
ICL adeptly diminishes the necessity for extensive datasets by leveraging a modest collection of examples, serving as illustrative prompts for both inputs and outputs. This approach not only enhances the models' adaptability but also significantly amplifies their proficiency in tackling tasks across a spectrum of novel and unseen instances. 
Heuristics are defined as $'$\textit{a high-level rule or strategy for inferring answers to a specific task.}$'$ and play a crucial role in human cognition. Humans use heuristics as an effective cognitive pathway, which often leads to more accurate reasoning than complex methods \cite{gigerenzer2011heuristic, Hogarth_Karelaia_2007, zhou2024llms}.
In ICL, heuristics are used to select or design examples (demonstrations) that can guide the model to make correct predictions\cite{zhou2024llms}. By using examples generated by the model itself as context, the reliance on large-scale training datasets can be reduced, enhancing the model's adaptability.
The performance of ICL is highly sensitive to specific settings, necessitating the selection of appropriate contextual information and the optimization of the model's training process. This includes the choice of prompt templates, the selection of context examples, and the order of examples \cite{pmlr-v139-zhao21c, lu-etal-2022-fantastically}, as well as the selection of examples and the format of inference steps  \cite{zhang2022automatic, fu2022complexity, zhang-etal-2022-active}, which collectively impact the application of ICL on LLMs.

The Chain-of-Thought (CoT) \cite{wei2022chain} stands as an augmented prompting technique, widely recognized for its efficacy across a spectrum of tasks that demand sophisticated reasoning. CoT has proven particularly adept at tackling complex reasoning challenges, encompassing arithmetic and commonsense reasoning \cite{cobbe2021training, wei2022chain}. However, its effectiveness is notably constrained in non-reasoning scenarios. When applied to tasks that do not inherently require reasoning, the CoT method risks simplifying the multi-step reasoning process into a potentially inadequate single-step, thereby undermining its full potential \cite{shum-etal-2023-automatic, zhou2024llms}.
Consequently, there is a compelling need to devise specialized prompting strategies tailored for non-reasoning tasks. These strategies should be crafted to address the unique demands of such tasks, ensuring that the models maintain their robust performance across the diverse landscape of language processing challenges.

In this paper, we introduce a suite of innovative contributions aimed at advancing Event Argument Extraction and addressing the limitations of existing methods:
\newline\textbf{Definition-augmented Heuristic-driven Prompting Method.} We improved the prompting heuristic method by incorporating argument extraction related definitions prompting and identified arguments. Utilizing inputs that include document content, task definitions, argument extraction rules, and identified event types and triggers, we constructed a definition-driven heuristic ICL. This method can process new situations (new classes) by analogy with known situations (known classes), effectively reducing error propagation and improving task accuracy.
It provides a structurally complete and well-defined framework for events and arguments, incorporating necessary constraints. This not only improves the precision of extraction but also offers the model a richer and more consistent reference benchmark.
\newline\textbf{Chain-of-Thought Method.} We employed the Chain-of-Thought method, guiding the model to incremental reasoning by providing coherent examples. These examples demonstrate how to break down complex problems into more manageable sub-problems and enhance the model's reasoning capabilities by simulating the human thought process.
\newline\textbf{Optimized Prompt Length.} For the document-level Event Argument Extraction task, we fine-tuned the prompt length to enhance overall extraction performance. Such adjustments ensure that the token limit of LLMs is not exceeded. The prompt contains sufficient information while avoiding efficiency decline due to excessive length.

We propose new perspectives and methods, solving the example selection problem from the new perspective of Definition-Enhanced Prompting Heuristic Method, promoting explicit heuristic learning in ICL. The aim is to build more robust and adaptable prompting methods suitable for Event Argument Extraction. By implementing proof, it effectively improves task performance, enhances the model's ability to grasp the complex relationships between events and arguments, and contributes to further improving the capabilities of LLMs in EAE tasks.

\section{Approach}
\begin{figure*}
  \includegraphics[width=\linewidth]{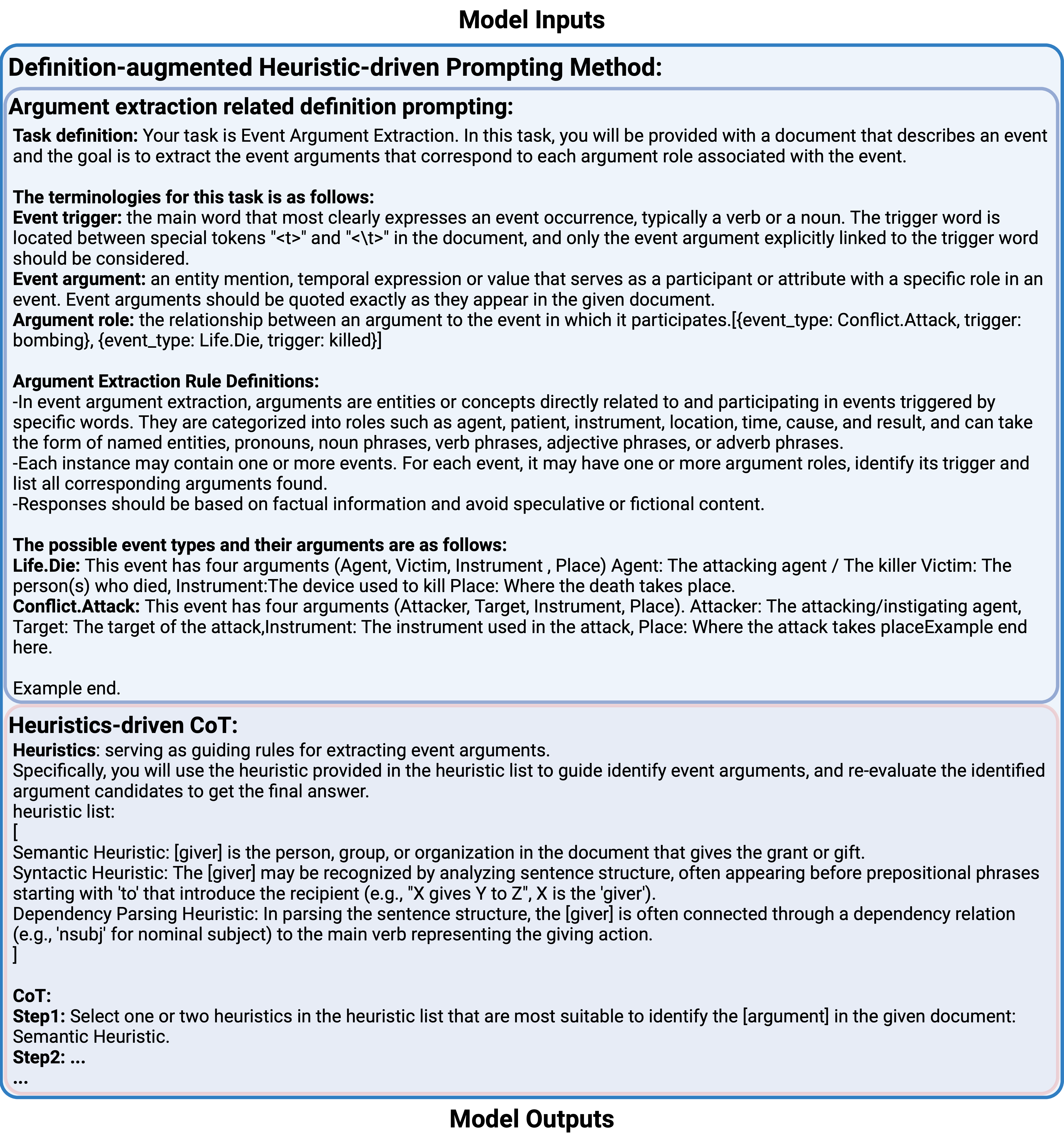}
  \caption{Definition-augmented Heuristic-driven Prompting method guides on how to extract event arguments related to specific trigger words from documents by defining the task, terminology, extraction rules, and a list of heuristics. It provides corresponding definitions for argument extraction prompting and heuristic rules to assist in the identification and extraction of event arguments.}
  \label{fig1}
\end{figure*}
We propose Definition-augmented Heuristic-driven Prompting Method for enhancing the performance of event argument extraction tasks. This method integrates argument extraction related definitions and rule-based knowledge, guiding the extraction process of event arguments through the introduction of heuristic rules, the main prompting process and content are illustrated in Figure \ref{fig1}.

The Argument extraction related definition prompting part mainly focuses on:
\newline\textbf{Event Attributes and Definitions:} Prior to argument extraction, it is essential to clarify the definition of events and associated terminologies. Events are defined as explicitly marked verbs or nouns in the document, with the verb or noun serving as the event trigger, and event arguments are entities, temporal expressions, or value concepts explicitly connected to this trigger, playing a certain role in the event. For instance, in an event defined as "Conflict.Attack," key event arguments include the attacker (Agent), victim (Victim), weapon (Instrument), and location (Place).
\newline\textbf{Argument Extraction Rules:} We employ a series of heuristic rules to guide the extraction of event arguments. These heuristic rules define potential argument roles based on the relationship between entities and event triggers, such as agents, patients, instruments, locations, times, outcomes, etc., and consider various morphological structures including noun phrases, pronouns, verb phrases, adjective phrases, and adverbial phrases.

The argument extraction related definitions serve as guiding rules for extracting arguments, assisting in the rapid identification of event arguments and reassessing them after identification to determine the final answers. We utilize semantic and dependency parsing heuristics, such as identifying the agent of an action and linking the agent to the verb through dependency relations, to enhance the identification of arguments.
Through these definitions, we are able to extract the trigger words for each event and all corresponding arguments, ensuring that the extracted information is fact-based and avoids speculative or fictional content.

For the Heuristics-driven CoT part, we mainly follow the settings and definitions proposed by \citet{zhou2024llms} and \citet{wei2022chain} to guide the model along a specific logical path, thereby improving the accuracy of event argument extraction. This leverages heuristic rules to inspire and guide the model through a logical chain from preliminary assumptions to final conclusions, revealing the complex structure and associations behind the event.
We have optimized parts of the reasoning process:
\newline\textbf{Initiation Phase:} The event triggers and potential arguments identified through Argument extraction related definition prompting initialize the starting point of the reasoning chain.
Reasoning Expansion: Based on heuristic rules, the model gradually expands the reasoning chain, parsing the potential relationships and attributes between event arguments through logical deduction. This phase emphasizes adding clear reasoning paths at each step to assist the model in more precise argument extraction in subsequent steps.
\newline\textbf{Logical Verification:} After the reasoning chain is preliminarily constructed, heuristic rules are used to logically verify the reasoning chain, ensuring the rigor of each step and adjusting potential logical errors.

Heuristic rules play a crucial role here, providing a logical foundation and directional guidance for the construction of the Chain-of-Thought. The definitions of these rules are based on an in-depth understanding and recognition of specific event types. For example, by analyzing the linguistic and semantic relationships between event triggers and potential arguments, the logical order and associations of these elements can be deduced.

Through the comprehensive application of these methods, our goal is to enhance the performance of event argument extraction tasks and strengthen the model's ability to grasp the complex relationships between events and arguments.

\section{Experiments}
\begin{table*}
\centering
\begin{tabular}{ccccc}
\hline
\multicolumn{2}{c}{\multirow{2}{*}{Method}} & \multicolumn{2}{c}{RAMS}   & DocEE-Normal       \\ \cline{3-5} 
\multicolumn{2}{c}{}                         & Arg-I & Arg-C & Arg-C \\ \hline
\multirow{5}{*}{Supervised-learning} & EEQA \citeyearpar{Du_Cardie_2020} & \multirow{5}{*}{-} & 19.54 & \multirow{4}{*}{-} \\
                                  & PAIE \citeyearpar{ma2022prompt}     &       & 29.86 &       \\
                                  & TSAR \citeyearpar{xu-etal-2022-two}      &       & 26.67 &       \\
                                  & CRP \citeyearpar{liu2023document}       &       & 30.09 &       \\
                                  & FewDocAE \citeyearpar{yang2023few}  &       & -     & 12.07 \\ \cline{1-5} 
\multirow{2}{*}{Llama3.1-70b}     & CoT \citeyearpar{wei2022chain}      & 39.80 & 30.69 & 26.11 \\
                                  & Ours     & \textbf{42.33} & \textbf{34.60} & \textbf{29.69} \\ \hline
\multirow{2}{*}{Deepseek-v2-chat} & CoT \citeyearpar{wei2022chain}     & 43.21 & 38.67 & 29.67 \\
                                  & Ours     & \textbf{48.00} & \textbf{45.54} & \textbf{31.33} \\ \hline
\end{tabular}
\caption{Overall performance. In few-shot setting, the scores of supervised learning methods on RAMS dataset are based on results reported in \citet{liu2023document}, where 1\% of the training data is used.}
\label{tab1}
\end{table*}

\subsection{Setup}
To evaluate the document-level Event Argument Extraction task, we adopt the RAMS \cite{Ebner_Xia_Culkin_Rawlins_VanDurme_2020} and DocEE \cite{tong2022docee} datasets. 
For the assessment, we follow the metrics outlined in \cite{ma2022prompt, zhou2024llms}, which are the F1 score for argument identification (Arg-I) and the F1 score for argument classification (ArgC). Detailed statistical data of the datasets and the number of test samples are listed in Appendix A.
Our Definition-augmented Heuristic-driven Prompting (DHP) method is compared with several state-of-the-art prompting methods, as well as the Chain-of-Thought (CoT) prompting \cite{wei2022chain}.

Here, we present the replication of results based on the CoT prompting method 
by \citet{zhou2024llms}, which represents one of the few excellent prompting strategies specifically tailored for the Event Argument Extraction task in LLMs. The experiments were conducted using two large language models: the publicly available Deepseek-v2-chat \cite{deepseekai2024deepseekv2strongeconomicalefficient} and Llama3.1-70b \cite{dubey2024llama}. It is noteworthy that due to the relatively high cost of Deepseek-v2-chat, its evaluation was limited to a subset of the dataset. Further experimental details can be found in Appendix A. We also have compared our approach with a variety of supervised learning methods found in the current literature. These include CRP \cite{liu2023document}, Few-DocAE \cite{yang2023few}, PAIE \cite{ma2022prompt}, TSAR \cite{xu-etal-2022-two}, and EEQA \cite{Du_Cardie_2020}. Within the domain of few-shot learning, our comparative analysis is grounded on the performance data from a limited number of samples as previously reported by \citet{liu2023document} and \citet{zhou2024llms}.

\subsection{Results}

\begin{table}
\small
\begin{tabular}{ccc}
\hline
\multicolumn{2}{c}{\multirow{2}{*}{Method}} & \multirow{2}{*}{DocEE-Cross} \\
\multicolumn{2}{c}{}                        &                              \\ \hline
Supervised-learning        & FewDocAE       & 10.51                        \\ \hline
Llama3.1-70b               & Ours           & \textbf{32.24}               \\ \hline
Deepseek-v2-chat           & Ours           & \textbf{33.43}   \\ \hline
\end{tabular}
\caption{In the cross-domain setting of the DocEE dataset, the Arg-C performance varies across different methods.}
\label{tab2}
\end{table}

Table \ref{tab1} presents experimental results that demonstrate our DHP prompting significantly enhances contextual learning for the document-level Event Argument Extraction (EAE) task. 

The DHP method consistently outperforms the CoT prompting \cite{wei2022chain} across LLMs and two datasets. Specifically, in the RAMS dataset, the DHP method achieves the largest F1 score improvements for Arg-I of 2.53\% and 4.79\%, and for Arg-C of 3.91\% and 6.87\%, respectively. Compared to supervised learning methods, the application of the DHP method in large models has led to Arg-C score improvements of 4.51\% and 15.45\%. This indicates that the DHP method significantly enhances the ability of large language models to identify arguments related to specific event triggers and assign them the correct argument roles.

In the DocEE dataset, under normal-setting, our method achieves substantial improvements over FewDocAE, with increases of 17.62\% and 19.26\%, respectively \cite{yang2023few}. The experimental results suggest that to further ascertain whether the DHP method can enhance the generalization capability of LLMs on data from different domains, which is crucial in real-world applications where large amounts of annotated data may be difficult to obtain \cite{tong2022docee, pmlr-v202-luo23e}, we tested the model performance under the Cross domain-settings of the DocEE dataset, as shown in Table \ref{tab2}. The large models with the DHP method also achieved at least a 21.73\% increase in the F1 score for Arg-C.

This supports the conclusion that our method can successfully reduce the reliance on large volumes of labeled data for document-level EAE tasks while improving accuracy.

\subsection{Analysis}
Following our empirical validation of the effectiveness of the DHP method, our approach naturally incorporates various heuristic methods into the prompts. By guiding the model to generate a detailed reasoning process, the accuracy and interpretability of the model are enhanced, which aids in more precisely identifying relationships between entities and improving the accuracy of argument extraction. We decompose the definitions related to the event argument extraction task to avoid performance degradation caused by handling too much information in a single task, thus overcoming the illusion problem.
The relevant prompting strategies applied by our DHP method can indeed effectively improve the LLMs performance of unseen classes in the prompts.

We believe that selecting appropriate models and configurations, coupled with carefully designed prompts and balanced datasets, is crucial for improving the performance of event extraction tasks. Moreover, cognitive research has found that compared to complex methods, humans use heuristics as an effective cognitive pathway to achieve more accurate reasoning \cite{gigerenzer2011heuristic, Hogarth_Karelaia_2007, zhou2024llms}. As similar results presented in the studies by \citet{wei2022chain} and \citet{zhou2024llms}, paralleling this human cognitive strategy, we enable LLMs to learn from explicit heuristics to enhance reasoning. Specifically, for LLMs that perform poorly under vague prompts and in non-reasoning tasks where it is difficult to grasp clear reasons, explicit heuristic specifications provide LLMs with a useful strategy for using and enhancing reasoning. By converting these implicit heuristics into explicit ones, a more direct way to utilize heuristics is provided, allowing LLMs to handle new situations by analogy with known cases. This capability is particularly useful in ICL, as LLMs are always faced with unseen samples and unseen classes \citep{zhou2024llms}.

\section{Related works}
\subsection{Document-level EAE}
Document-level EAE commonly relies on manual domain and pattern annotation for supervised learning models \cite{xiang2019survey, Lin_Ji_Huang_Wu_2020, li2022survey, liu-etal-2022-dynamic, hsu2022degree, liu2023document}.
The high costs, coupled with the reliance on extensive manually annotated data, may pose a bottleneck for their practical application \cite{lin-etal-2023-global}. \cite{agrawal-etal-2022-large} have employed LLMs in clinical Event Argument Extraction (EAE) using standard prompts that do not involve any reasoning strategies, while research on prompting strategies specifically tailored for the EAE task is scarce, with only \cite{zhou2024llms} exploring the promising and challenging research direction of reducing the dependence on specific large-scale training datasets through ICL, thereby enhancing the generalization capability of LLMs in EAE tasks.
\subsection{In-Context Learning}

The In-Context Learning (ICL) \cite{brown2020language} methodology is designed to expedite the adaptability of language models across various tasks, necessitating minimal or no prior data \cite{wei2022chain, kojima2022large}. This methodology eschews direct fine-tuning through the capacity for models to interpret and perform tasks drawing on contextual clues. \citet{weber2023icl} enhanced model accuracy by employing carefully crafted efficient prompting templates and diverse prompting formats.
\citet{gonen2023demystifying} have noted that the performance of ICL 
is highly sensitive to the selection of examples. \citet{zhou2024llms} innovatively explored the use of examples to guide Large Language Models (LLMs) in processing specific tasks through heuristic rules. 
This implies that well-designed prompts and heuristic rules can effectively enhance ICL performance without the need for fine-tuning on task-specific datasets.

\section{Conclusion}
In this study, we propose a Definition-augmented Heuristic-driven prompting strategy for LLMs in document-level event argument extraction tasks. 
Through experimentation, we have found that incorporating Argument Extraction Related Definition prompting can further enhance the performance of event argument extraction, building upon structured heuristic methods and the Chain-of-Thought approach. Our method has exhibited consistent performance and generalization capabilities across two datasets, showing potential and application prospects.

\bibliography{custom}

\appendix

\section{Experimental Details}
\label{sec:appendix}
\begin{table}
\begin{tabular}{cccc}
\hline
Dataset & \# Example & \# Eval. & Eval. Split \\ \hline
RAMS \shortcite{tong2022docee}    & 1          & 871      & test        \\ \hline
DocEE \shortcite{Ebner_Xia_Culkin_Rawlins_VanDurme_2020}  & 1          & 800      & test        \\ \hline
\end{tabular}
\caption{The overall statistics of the dataset. \# Example: The number of examples used in the HDP method. \# EVAL.: the number of samples used for evaluation of different prompting methods. EVAL. Split: evaluation split.}
\label{tab3}
\end{table}

The dataset statistics are presented in Table \ref{tab3}.
For the large scale of the DocEE and RAMS datasets, full-size evaluation using LLMs is impractical. We follow the setup of \citet{shum-etal-2023-automatic, wang2022rationale, zhou2024llms}, Wang et al. (2022), and Zhou et al. (2024), and evaluate a subset of these datasets. Due to the substantial costs associated with deploying LLMs, we limit our assessment to 200 samples for both the RAMS and DocEE datasets.
Furthermore, for the DocEE dataset, it presents two distinct settings. In the conventional configuration, the training and testing data share an identical distribution. Conversely, the cross-domain setup features training and testing data composed of non-overlapping event types \cite{tong2022docee,zhou2024llms}.
\end{document}